\documentclass[manuscript]{acmart}

\AtBeginDocument{%
  \providecommand\BibTeX{{%
    \normalfont B\kern-0.5em{\scshape i\kern-0.25em b}\kern-0.8em\TeX}}}

\setcopyright{acmcopyright}
\copyrightyear{2018}
\acmYear{2018}
\acmDOI{10.1145/1122445.1122456}

\acmConference[ETRA '21]{ETRA '21: ACM Symposium on Eye Tracking 
	Research & Applications }{June 03--05, 2018}{Stuttgart, BW}
\acmBooktitle{Woodstock '18: ACM Symposium on Neural Gaze Detection,
  June 03--05, 2018, Woodstock, NY}
\acmPrice{15.00}
\acmISBN{978-1-4503-XXXX-X/18/06}

\begin{document}

\title{Differentiating Surgeon Expertise Solely by Eye Movement Features}

\author{Benedikt Hosp}
\email{benedikt.hosp@uni-tuebingen.de}
\orcid{https://orcid.org/0000-0001-8259-5463}
\affiliation{
  \institution{Human-Computer Interaction, University of Tübingen}
 \streetaddress{Sand 14}
  \city{Tübingen}
  \country{Germany}
  \postcode{72076}
}

%\author{Name}
%\%affiliation{
%	\institution{University of XXX}
%	\streetaddress{xxx Street x}
%	\city{xxx}
%	\country{xxx}}
%\email{xx@xxx.org}

\author{Myat Su Yin}
\affiliation{%
 \institution{Faculty of ICT, Mahidol University}
\streetaddress{}
\city{Nakhon Pathom}
  \country{Thailand}}
\email{myatsu.yin@mahidol.ac.th}

\author{Peter Haddawy}
\affiliation{%
	\institution{Faculty of ICT, Mahidol University}
	\streetaddress{}
	\city{Nakon Pathom}
	\country{Thailand}}
\email{peter.had@mahidol.ac.th}

\author{Paphon Sa-Ngasoongsong}
\affiliation{%
	\institution{Faculty of Medicine, Ramathibodi Hospital, Mahidol University}
	\streetaddress{}
	\city{Bangkok}
	\country{Thailand}
\email{paphonortho@gmail.com}}

\author{Enkelejda Kasneci}
\orcid{https://orcid.org/0000-0001-8259-5463}
\affiliation{
	\institution{Human-Computer Interaction, University of Tübingen}
	\streetaddress{Sand 14}
	\city{Tübingen}
	\country{Germany}
	\postcode{72076}
}

\renewcommand{\shortauthors}{Hosp et al.}%{Hosp et al.}

\begin{abstract}
%Enwicklungen der Informatik der letzten Jahre ziehen auch in die Krankenhäuser ein. Chirurgen sehen sich immer neuen technischen Herausforderungen ausgelioefert. Die visuelle Wahrnehmung spielt dabei eine Schlüsselrolle. Um das Training von jungen Chirurgen zu optimieren bedarf es Diagnostic und Trainingsmodelle. In dieser Studie stellen wir ein Model zur Klassifizierung von Experten, Residents im 4. Jahr und Reseidents im 3rd Jahr, nur über die Augenbewegungen vor. Wir zeigen ein model, dass mit einer minimalen Featureauswahl auskommt und dennoch eine Robuste Accuracy von 76.46 \% die Augenbewegungen in die korrekte klasse zuordnen kann. Ebenso gehen wir in dieser Studie auf die Entwicklungsschritte der visuellen Wahrnehmung zwischen drei expertise Klassen ein und bilden damit einen ersten Schritt in Richtung Diagnosemodel für Expertise.

Developments in computer science in recent years are moving into hospitals. Surgeons are faced with ever new technical challenges. Visual perception plays a key role in most of these. Diagnostic and training models are needed to optimize the training of young surgeons. In this study, we present a model for classifying experts, 4th year residents and 3rd year residents, using only eye movements. We show a model that uses a minimal set of features and still achieve a robust accuracy of 76.46 \% to classify eye movements into the correct class. Likewise, in this study, we address the evolutionary steps of visual perception between three expertise classes, forming a first step towards a diagnostic model for expertise.

\end{abstract}




\maketitle

\section{Introduction}

Arthroscopy is a popular procedure that promises better patient outcomes on the one hand and conserve hospital resources on the other.  According to Monson et al. \cite{monson1993advanced}, patients experience less pain, have fewer complications and recover faster. However, a surgeon needs advanced technical skills for this type of operation (Hermens et al. \cite{hermens2013eye}), for example a technical understanding and adequate conception of the 3D tissue in a 2D view and the separation of the field of view (screen) and the instrument plane. This separation presents a challenge insofar as cognitive projections of motion from instruments in 3D space must be made into the 2D space of the screen. Likewise, the instruments determine and constrain the degrees of freedom of movement. Therefore, in order to reach the appropriate site during an operation via arthroscopy, navigation plays a central role for the surgeon. This involves inserting a camera through a portal into the patient's shoulder. The camera can rotate in multiple dimensions and casts its image on a screen placed next to the patient, so that surgeons largely rely on this image during surgery. The complex anatomy as well as the difficult perception due to rotation of the camera into 3D tissue, pose a challenge. Due to these technical challenges, there is a growing interest within the scientific community to optimize training. In particular,the role of eye movements is increasingly being investigated. Therefore, diagnostic systems are needed to improve navigation via perception, especially for non-experts and to understand the perception of surgeons during surgery and despite that to draw conclusions for training. In order to find out whether a perceptual-cognitive training system is useful, first it must be determined whether, and to what extent, differences in surgeons' expertise are reflected by their eye movements. The findings from this diagnostic are significant for the design of adequate training scenarios for perceptual-cognitive diagnostic and training systems.

%\subsection{What is the goal?}

In this work we consider the perception of surgeons using eye movement patterns from three expertise levels during an arthroscopic surgery of the shoulder. We use stimulus-independent eye movement patterns to classify the expertise of the subjects and to investigate differences in the eye movement behavior of the subjects in detail. These findings are not only useful for training, but also for assigning surgeons with different skills to different complex operations.

%Was gibt es für ähnliche Arbeiten?

% Verbindung aus drei Dingen: ET, ML und Chirurgie/Arhtroskopie
%-> ET in Chirurgie/Arthoscopie
%-> Klassifikation von ET Daten in Expertise
%-> ML zur Analyse

%\subsection{Realism vs. artificial}
A fundamental aspect of perception research is the validity of the presentation as a highly natural presentation allows the participants to apply their natural gaze behaviour. Many studies were conducted using artificial forms of presentation like virtual reality (VR)~\cite{law2004eye,zheng2011surgeon,wilson2011perceptual,atkins2012saccadic} or images \cite{sodergren2010hidden,eivazi2012gaze}. While this kind of presentation allows a certain level of control for the experimenter, artificially created presentations are hiding important influences the participant needs to infer. Zheng et al. \cite{zheng2011surgeon} for instance report higher levels of frustration because the physical demands in their study were low, meaning their simulator might prevent optimal settings for expertise research.  To provide a presentation mode that is as natural as possible, we use so called soft cadavers that provide natural tactile sensation and therefore increase the naturality of the scene. Another important aspect in perception research is the level of control. 

%\subsection{ Experimental control: Head vs remote}
In lab studies, remote eye trackers are usually used ~\cite{law2004eye,atkins2013surgeons}, since remote eye trackers typically have a screen as stimulus output and collect data in relation to this input screen, a comparison between participants is much easier as the control about the stimulus is totally in the experimenters hands. A week point of remote eye trackers is their flexibility. As soon as the participant is looking into another direction (i.e. down at the cadaver), remote eye trackers cannot capture the gaze signal anymore. To allow the participant to apply a normal gaze behaviour and move freely without data lost, we use a head-mounted eye tracker in combination with a 4k-screen. This setup supports natural gaze behaviour as well as high control on the stimulus. Thus, we can capture highly detailed information of the tissue on a screen with high resolution and gaze signals on the cadaver, both with the same field camera.

%\subsection{ET features}
%ET feature and expertise
As eye trackers provide a direct measuring method, they generally allow a high experimental control. Eye trackers, in fact, provide a sound foundation with high temporal as well as spatial resolution to research on perceptual processes. However, studies in surgery also differ in how they evaluated the gaze signal. The gaze signal on the stimulus was considered, i.e. target gaze behavior, switching behavior (alternating gaze between target and instrument), or following behaviour (eye following the instrument) \cite{law2004eye}. Other studies focused on quiet eye periods \cite{wilson2011perceptual}. However, there are also studies that have gained insights at feature level. For example, Kocak et al.\cite{kocak2005eye} used stimulus-independent eye features in their binary classification and found significantly lower saccade rates as well as significantly higher peak velocities for experts, which was confirmed by other studies, e.g., by Hermens et al.\cite{hermens2013eye}. Tien et al. \cite{tien2011quantifying} found a higher fixation rate in experts. Richstone et al. \cite{richstone2010eye} approved a binary classification with a mixture of fixation rates, vergence, blink rates and pupil diameter variability. Eivazi et al. \cite{eivazi2012gaze} show differences in time to first fixation and mean fixation duration. However, theses differences were not approved by Sondergren et al. \cite{sodergren2010hidden}, as in both studies fixation durations are analyzed differently and the choice of regions of interest plays an important role. These results show that eye movements can be used to assess the surgical expertise and to define differences between groups. A wide range of eye tracking features has been shown to reflect expertise.

%\subsection{expertise}

% bisher nur experte und nicht experte, aer keiner zeigt wie sich die Sicht entwickelt
%UMSCHREIBEN
%Viele Studien konzentrieren sich dabei auf die Erkennung von Unterschieden in der Expertise zwischen Experten und Novizen \cite{wilson2011perceptual,zheng2011surgeon,tien2010measuring,tien2011quantifying,schulz2013situation}. Wenige Studien haben sich dabei mit der Entwicklung der Augenbewegungen beschäftigt. Studien mit dem Focus auf der Entwicklung nutzten jedoch simulatoren \cite{kocak2005eye}. Durch die betrachtung von verschiedenen Expertise stufen und der ausprägungen deren augenbewegungen erhoffen wir uns über die entwicklung des blickverhaltens durch die veränderung und unterschiede von augenbewegungen aussagen treffen zu können.

In this regard, many studies have focused on the detection of differences in expertise between experts and novices \cite{wilson2011perceptual,zheng2011surgeon,tien2010measuring,tien2011quantifying,schulz2013situation} and only few studies have focused on the development of eye movements. However, studies focusing on development used mostly simulators \cite{kocak2005eye} or images \cite{sodergren2010hidden}. By looking at different expertise stages and the characteristics of their eye movements, we seek to make statements about the development of gaze behavior through the changes and differences in eye movements of different classes. We only consider relative features and avoid absolute features like operation length as such can distort the picture. Thus,  we use only eye movement features to classify perception in isolation and apply the same methods like Hosp et al. \cite{plosoneHospGoalkeeper} in their study with athletes.

%\subsection{Machine learning}

Machine learning is the state-of-the-art when it comes to big data. As traditional methods quickly reach their limits when complex data are involved, machine learning methods are increasingly finding their way into the analysis of eye tracking data. Another point is that machine learning techniques are data-driven and therefore can learn patterns from previous data. Usually, the more data there are available for training, the more robust is the result. So far, several algorithms have been introduced to eye tracking including supervised methods like support vector machines \cite{tong2001support,plosoneHospGoalkeeper,castner2018scanpath} or neural networks \cite{castner2020deep}. However, hidden markov models (HMM) have also been used for this purpose. In the study by Sodergren et al. \cite{sodergren2010hidden} it was shown that their HMM can reveal differences in eye movement patterns between high and low performers. Ahmidi et al. \cite{ahmidi2010surgical} mixed instrument movements with eye movement data and achieved a binary classification accuracy of 82.5\% for skill level classification and confirmed their result with a second study \cite{ahmidi2012objective}. All these studies show that eye movement data can be used to differentiate between experts and novices and that it is not necessary to determine where exactly the surgeons were looking at to measure their skill accurately.

%Abgrenzung zu bisherigem Paper von Lin , Peter und mir
This work is based on the data of Yin et al. \cite{yin2020study}, in the sense that the same eye tracking data are used for our classification. In contrast, however, the evolution of eye movement patterns is done over  three classes, namely, novice (residents, 3rd year), intermediates (residents,4th year) and experts (fellow surgeons). In addition, we look at the characteristics of the features in order to obtain statements about the perception of the different groups. Since we want to look at perception in isolation and draw conclusions about expertise, our evaluation is completely limited to the use of relative eye movement features. No other features like operation length, number of saccades/fixations etc. are used.
\section{Participants and Methods}

\subsection{Procedure}

\begin{figure*}
\centering
\includegraphics[width=0.5\linewidth]{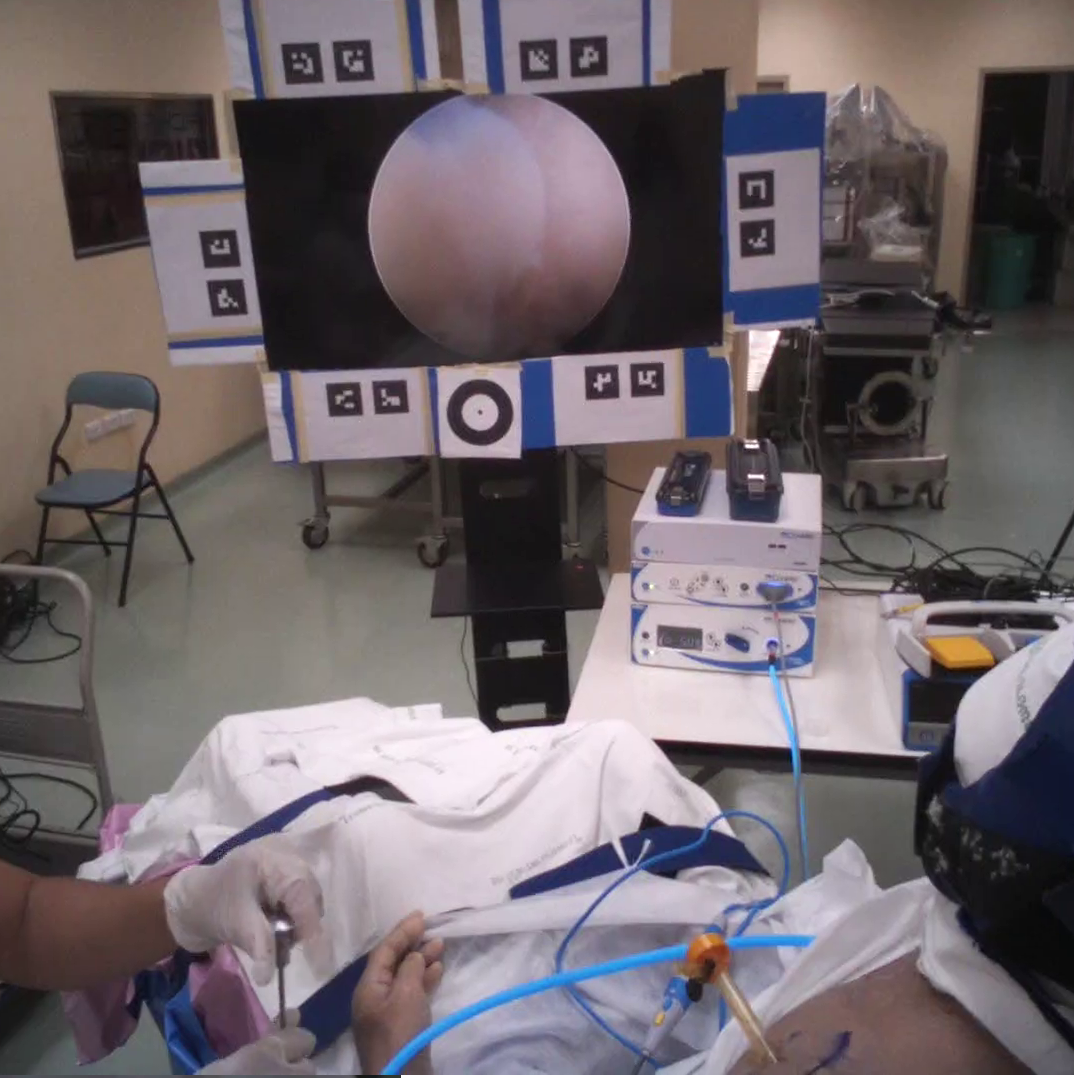}
\caption{Study setup with 4k-screen four feet away from the surgeon and cadaver directly in front of surgeon. }~\label{fig:setup}
\end{figure*}

The data of this study belong to participants from three different groups of surgeons. The expert group contains n = 5 fellow surgeons with 4-10 years of experience from the Orthopaedics Faculty of Medicine, Mahidol University. N=5 participants are in the 4th year of the Ortopaedic Surgery Residency Program and belong to the intermediates group. The novice group contains n=5 participants, who are in the 3rd year of the same program. All residents have no arthroscopic experience. The study was approved by the Mahidol Unversity Institutional Review Board. Participants were placed in front of the cadaver and four feet away from the 4k, 52-inch screen where the output of the arthroscope is displayed (see Figure \ref{fig:setup} ). Each participant was familiarized with the setup and asked to navigate to 12 anatomical landmarks in the shoulder, while having a Tobii Pro Glasses 2 eye tacker on. The gaze was recorded with Tobii software.

\subsection{Data preparation}
\label{dataPrep}

The Tobii Glasses 2 were set to a frame rate of 100 Hz. This means a gaze sample should be available every 10 ms and is saved with a timestamp, x- ,and y-coordinates. These samples are used to calculate metrics like fixations and saccades. To conduct this process we used the Tobii Fixation Filter, which behaves mainly like a I-VT filter (velocity threshold based) but uses a sliding window averaging method. The feature calculation is based on the classification algorithm of Olson \cite{olsson2007real}. We used the default values, which result in a velocity threshold of 0.7 pixels/ms. The raw eye tracking data, as well as the fixations and saccade metrics, can be exported from the Tobii Studio software. As our aim is to infer which features contribute to expertise differences, we first use all the exported features from Tobii Studio Software and added common metrics, too. Subsequently, we evaluated their frequency in the model building process and rated the most frequent features to build a model with this subset of features for expertise acquisition. For this classification we focus first on the following features:

\begin{itemize}
	\item Velocity of saccades (average, min, max. std. dev.)
	\item Amplitude of saccades (average, min, max. std. dev.)
	\item Total amplitude of saccades
\end{itemize}

From the velocity of the saccades, as well as the amplitude of saccades, we use the person-specific average, minimum, maximum and standard deviation as features. Next to the total amplitude of saccades Tobii provides metrics about the first saccade and first fixation. We did not include them since our participants have been familiarized with the glasses in different length and therefore we end up with chaotic first saccades, which have no informative character, when the trial starts. 

We added certain typical eye movement features which we calculated by ourselves.

\begin{itemize}
	\item Saccade duration (average, min, max. std. dev.)
	\item Fixation duration (average, min, max. std. dev.)
	\item Fixation frequency
	\item Saccade frequency
	\item Pupil diameter (average, min, max. std. dev.)
	\item Gyroscope X,Y,Z (average, min, max. std. dev.)
\end{itemize}

We decided to include the gyroscope values as we consider a lot of information about head movement between screen and cadaver may be revealed. The integration of pupil diameter features is based on the assumption, that experts may have less fluctuating puil diameter since there mental effort is considered to be smaller. Vice versa, the pupil diameter of intermediates and novices may reveal expertise differences by such effects.
Our first model is built on these mentioned 35 features.

\subsection{Machine learning model}

We took all 34 features and built a SVM model. We protected the model against overfitting by partitioning the data set into folds and estimate the accuracy on each fold by applying a 12-fold cross-validation. On each of 1000 runs we kept out one participant of each group. These participants have never been seen by the model before (leave-one out validation). Therefore, we trained the model with the remaining four participants of each group and evaluated each run with the three participants of the leave-one out validation data set. The membership of a participant to validation or training set is decided randomly in each run.

%\subsubsection{Model building}
%To find a model robust to high data variations, we applied a cross-validation during training. The final model is based on the average of k=50 models, with k = number of folds in the cross-validation. For each model $m_i$, with $i \in \{1, \dots ,k\}$, we use all out-of fold data of the i-th fold to train and evaluate $m_i$ with the in-fold data of the i-th fold (this procedure is illustrated in Fig ~\ref{fig:trees}). The final model is evaluated with a leave-out validation. The cross-validation step during training is independent from the leave-out validation with totally new data (never seen by the model). Information from cross-validation is used during the building and optimizing of the model and leave-out validation solely provides information about the prediction accuracy of the model when using completely new data. 

%\subsection{Classifiability}
Firstly, we used all 34 features to check the classifiability of the data set and afterwards reduced the amount by taking the 4 most frequent features of 1000 runs. The most frequent features are features that have the highest importance values for a single model prediction. In each run we built a queue of all 34 features sorted by importance for the current model. Subsequently, we computed their overall frequency over all models.

\section{Results}

\begin{table}[!h]
	\centering	
	\caption{{\bf The most important and frequent features on 1000 runs.}}
	\begin{tabular}{|c|c|}
		\hline
		Feature & derivation\\
		\hline		 		
		Peak velocity of saccades & standard deviation\\	\hline		 		
		Amplitude of saccades & minimum \\\hline
		Total amplitude of saccades & sum\\\hline
		Gyroscope z & min \\\hline
		\hline
	\end{tabular}
	\label{tbl:mff}
	\begin{flushleft}\end{flushleft}		
\end{table}

Our first classifiability model shows promising results with an average accuracy of 60\%.
As a system that would simply guess the class, would only reach a chance-level of 33.33\%, the all feature model can already be considered as well-performing. But as we want to specify the results to allow a precise statements about a high performing classification with the least amount of features, we continued by collecting all features and their importance values, shown in table \ref{tbl:mff}, on 1000 runs of the all feature model. With a subset of 4 features, earlier counteracting features may be avoided and a precise statement about the differences of the groups can be stated.

\subsection{Performance metrics}

% Accuracy
% Precision
% Recall
% miss rate
% f1-score

\begin{figure*}
	\centering
	\includegraphics[width=1\linewidth]{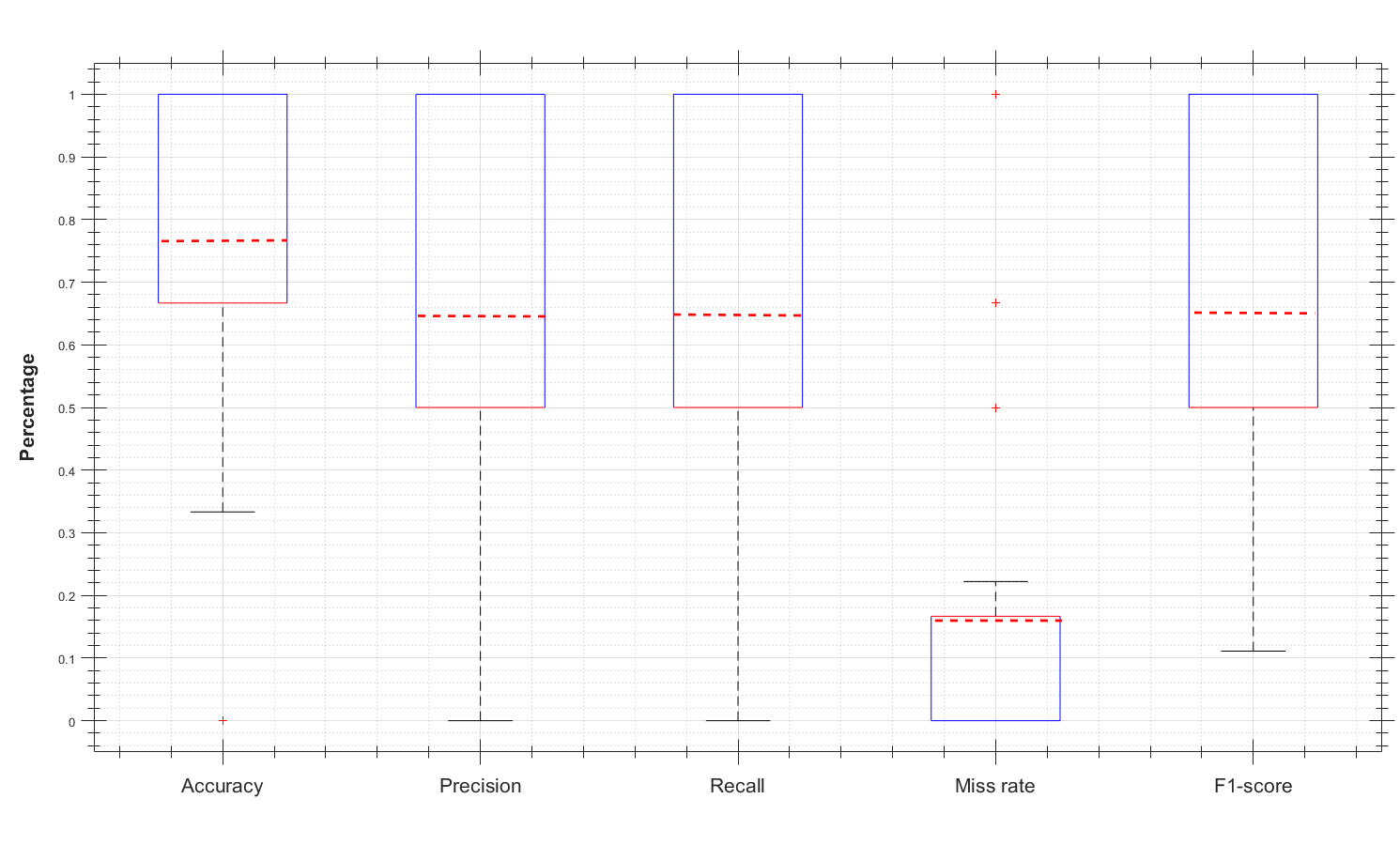}
	\caption{Performance values on 1000 runs. }~\label{fig:all}
\end{figure*}

Boxplot in Figure  \ref{fig:all} shows in the first boxplot to the left, the distribution of the accuracy values on 1000 runs. In 430 runs out of 1000 the model reached an accuracy of 90-100\%. The accuracy of 446 out of 1000 runs lays between 60-70 \%. Only 112 out of 1000 runs show accuracy values of 30-40\% and 12 out of 1000 runs show 0-12\%. This means that 98.8\% of the runs were better than chance level and 87.7\% still better than twice the chance-level. The dashed red line shows the average accuracy of 76.46\% of the model, based on the 4 most frequent features. This is a solid value for a three class classification. The average accuracy of the model is more than twice the chance-level.
As the accuracy of the model is a more general information provider, we calculated other metrics, too.
To know about correct predictions, we calculated the precision or positive predictive value, shown in Figure \ref{fig:all}, second to left. The precision indicates the proportion of results correctly classified as positive in the totality of results classified as positive. It is a metric that tells about how precise the predictions of the model are. In other words, the probability of a sample being predicted to belong to class x which really belongs to class x. Out of 1000 runs, 430 runs had a precision of 100\%. Only 12 of the runs were between 0-10\%. The same behaviour can be seen for the recall (Figure \ref{fig:all}). Recall, or hit rate, is a measure that tells about the probability that a positive object is correctly classified as positive. Recall is especially important as a model metric, when there are high costs associated with false negatives. The fourth boxplot to the left shows the miss rate of the model \ref{fig:all}. Out of 1000 runs, the model had a miss rate between 0.05-0.2\% on 480 runs. On 380 runs the miss rate was even smaller at 0.0-0.05\%. Accordingly, the false negative rate or miss rate indicates the proportion of objects falsely classified as negative out of the total number of positive objects. The fifth boxplot in Figure \ref{fig:all} shows F1-scores. With f1-scores we have a metric that balances between precision and recall. The F1-score combines precision and recall by means of the weighted harmonic mean, where precision and recall are weighted equally. The performance of the model is balanced well, as can be seen in the boxplot. The average f1-score is at 50\%. 480 out of 1000 runs achieved a f1-score of 50\%-60\%. Another 380 runs achieved even 90\%-100\%.

\subsection{Feature evolution}

As we can see, that the model is performing very well and therefore able to classify the three groups of expertise, we have a look at the feature characteristics as well to analyze their evolution. We do that with the 4 most frequent features from Table \ref{tbl:mff}. Table \ref{tbl:evolution} contains the characteristics of the 4 MFF features. The table shows that experts have a smaller standard deviation of the peak velocity of the saccades (93.26 \textdegree/s). This feature is hard to interpret, but one assumption may be, that experts have a more uniformly distribution of saccade velocites. This means they do more saccades at the same speed, compared to intermediates and novices. Interestingly, intermediates as the middle class between expert and novice show a much more divers saccade peak velocity behaviour (121.72 \textdegree/s). Novices are in the middle between experts and intermediates. A higher value for the standard deviation of the saccade peak velocities could a an indicator for a more chaotic gaze behaviour, but it is hard to draw a conclusion on such feature. When having a look at the minimum saccade amplitudes, we can see the same differences. The experts have on average a bigger minimal saccade of the length of 0.86 \textdegree, compared to the intermediates with 0.40\textdegree and the novices with 0.64\textdegree. Again, we can see that the novices are in between the experts and intermediates. Only the total amplitude of all saccades show a uniform evolution. The experts do a total of 481.32\textdegree of saccade length, where intermediates do more than twice the experts (1120.74\textdegree) and novices (1956.21\textdegree) even more than five time the experts and nearly double the intermediates. Another interesting feature evolution can be seen at the gyroscope minimum measure in z-axis. This corresponds to head rotations to left and right. With -66.47\textdegree the intermediated have the smaller average value of z-axis rotation, followed by the experts with -72.90\textdegree and the novices with -80.12 \textdegree.

\begin{table}[!h]
	\centering	
	\caption{{\bf Average feature evolution between classes.}}
	\begin{tabular}{l| c |c |c}
%		\hline
		Feature & Expert(Fellow) & Intermediate (R4) & Novice (R3)\\
		\hline		 		
		Sasccade peak velocity (std. dev.) & 93.26 \textdegree/s & 121.72 \textdegree / s & 117.45 \textdegree/s \\ \hline
		Saccade amplitude (min) & 0.86 \textdegree & 0.40 \textdegree & 0.64 \textdegree \\ \hline
		Total saccade amplitude & 481.32 \textdegree & 1120.74 \textdegree & 1956.21 \textdegree \\ \hline 
		Gyroscope Z (min) & -72.90 \textdegree & -66.47 \textdegree & -80.12 \textdegree \\ %\hline
		
%		\hline
	\end{tabular}
	\label{tbl:evolution}
	\begin{flushleft}\end{flushleft}		
\end{table}

\section{Discussion}

In this work we developed a model with supervised machine learning techniques, that is able to distinguish three levels of expertise solely on eye movements during an arthroscopic surgery of the shoulder. With an accuracy of 76.46\% the model can be considered as performing well, since it is two times better than guessing. Even for a binary classification, this result would be high. We showed that different performance metrics (accuracy, precision, recall, miss rate and f1-score) allow to state that the model is robust, precise and valuable to be used in a training system. To further understand the differences between the three groups of expertise, we had a look at the 4 most frequent features of the model and analyzed the evolution of the characteristics between the groups. Except for the total amount of saccade amplitudes, the three of the four most frequent features show an uniform evolution. First, novices tend to have a more chaotic gaze behaviour and distribute their gaze over a bigger amount of the scene by making lots of different saccades with different speed. They also tend to look more on the outside than on the center and move their head more from left to right and vice versa, which might indicate a problem in the cognitive projection of the 2D view of the arthroscope into the 3D-tissue in front of them. The evolution to intermediates shows an atypical behaviour, as they tend to still gaze over a bigger area of the scene than the experts, but do smaller saccades with a still divers velocity. They move their head less. This might indicate, that they try to focus on more specific visual clues and start to concentrate on the center of the scene. In the next evolution step, the saccade velocities shrink significantly, which signifies a more planned scanning behaviour, with little longer saccades, concentrated more on specific areas but with more head movements to the left or right. To summarize our findings, one can state that the evolution of novices to intermediates first tend to lead to a partly more chaotic gaze behaviour and then turning to be more precise. 

Further steps are to add more participants to the single classes, and refine the number of classes. This allows a much finer classification and therefore a better understanding of the differences between the classes. A finer classification is important to robust assumptions made by the model about gaze behaviour. In future we plan to add more participants from more classes to investigate on expertise differences in eye movements.

\bibliographystyle{ACM-Reference-Format}
\bibliography{acmart}

\end{document}